\documentclass[lettersize,journal]{IEEEtran}
\usepackage{amsmath,amsfonts}
\usepackage{algorithmic}
\usepackage{algorithm}
\usepackage{array}
\usepackage[caption=false,font=normalsize,labelfont=sf,textfont=sf]{subfig}
\usepackage{textcomp}
\usepackage{stfloats}
\usepackage{url}
\usepackage{verbatim}
\usepackage{graphicx}
\usepackage{cite}
\usepackage{booktabs}
\usepackage{multirow}
\usepackage{caption}
\usepackage{xcolor}
\usepackage[bookmarks=true]{hyperref}
\title{\texorpdfstring{\acceptbanner}{}SEVI: Safe Evidential Real-World Reinforcement Learning}

\author{Shaekh Mohammad Shithil and Lantao Liu%
\thanks{%
The authors are with the Luddy School of Informatics, Computing, and
Engineering, Indiana University, Bloomington, IN 47408 USA
(e-mail: \texttt{\{sshithil, lantao\}@iu.edu}).%
}%
}

\newcommand{\acceptbanner}{%
  {\normalfont\small\color{gray}%
    This paper has been accepted for publication in
    \textbf{IEEE Robotics and Automation Letters}
    \textcolor{magenta}{\href{https://ieeexplore.ieee.org/xpl/RecentIssue.jsp?punumber=7083369}{(RA-L)}},
    September 2026.\par}%
  \vspace{14pt}%
}

\begin{document}
\maketitle
\begin{abstract}
Real-world reinforcement learning has enabled contact-rich manipulation tasks to be trained directly on physical robots, but unsafe interactions during online learning remain a key barrier to practical deployment. We propose Safe Evidential Real-World Reinforcement Learning (SEVI), a framework for reliability-aware safety intervention during real-world policy learning. SEVI introduces an Evidential Safety Critic that estimates both future constraint-violation risk and epistemic uncertainty in a single forward pass. During online learning, SEVI evaluates each task-policy action before execution; if the action is predicted to be unsafe or the prediction is insufficiently supported by evidence, SEVI overrides it with an offline-trained recovery policy that guides the robot toward a lower-risk region. We evaluate SEVI on two simulated manipulation tasks and a real-world USB insertion task under in-distribution and distribution-shifted conditions. Across task-learning and safety-critic baselines, SEVI achieves the strongest trajectory-level safety prediction, the highest final task success, and the lowest unsafe-event counts. SEVI also reduces unsafe events by 13.4--57.1\% relative to recovery-based safe RL while preserving low-latency execution-time filtering. The project page is available at:
\href{https://sevi-safe.github.io/}{\texttt{https://sevi-safe.github.io/}}.
\end{abstract}
\vspace{-2 pt}

\section{Introduction}
Reinforcement learning (RL) holds the promise of enabling robots to acquire complex manipulation skills through trial and error. Recent real-world RL systems have shown that contact-rich manipulation tasks can be trained directly on hardware within a practical time budget. These systems typically rely on demonstrations, sample-efficient off-policy learning, learned rewards, and carefully designed controllers \cite{ball2023efficientonlinereinforcementlearning,luo2024serl}. 
However, training RL on physical robots remains challenging because exploration can lead to unsafe interactions, such as excessive contact forces, object damage, workspace violations, or failures requiring human intervention. Therefore, as real-world RL becomes increasingly sample-efficient, a central remaining challenge is to reduce unsafe interactions during online training.
\begin{figure}[t]
    \centering
    \includegraphics[width=0.95\columnwidth]{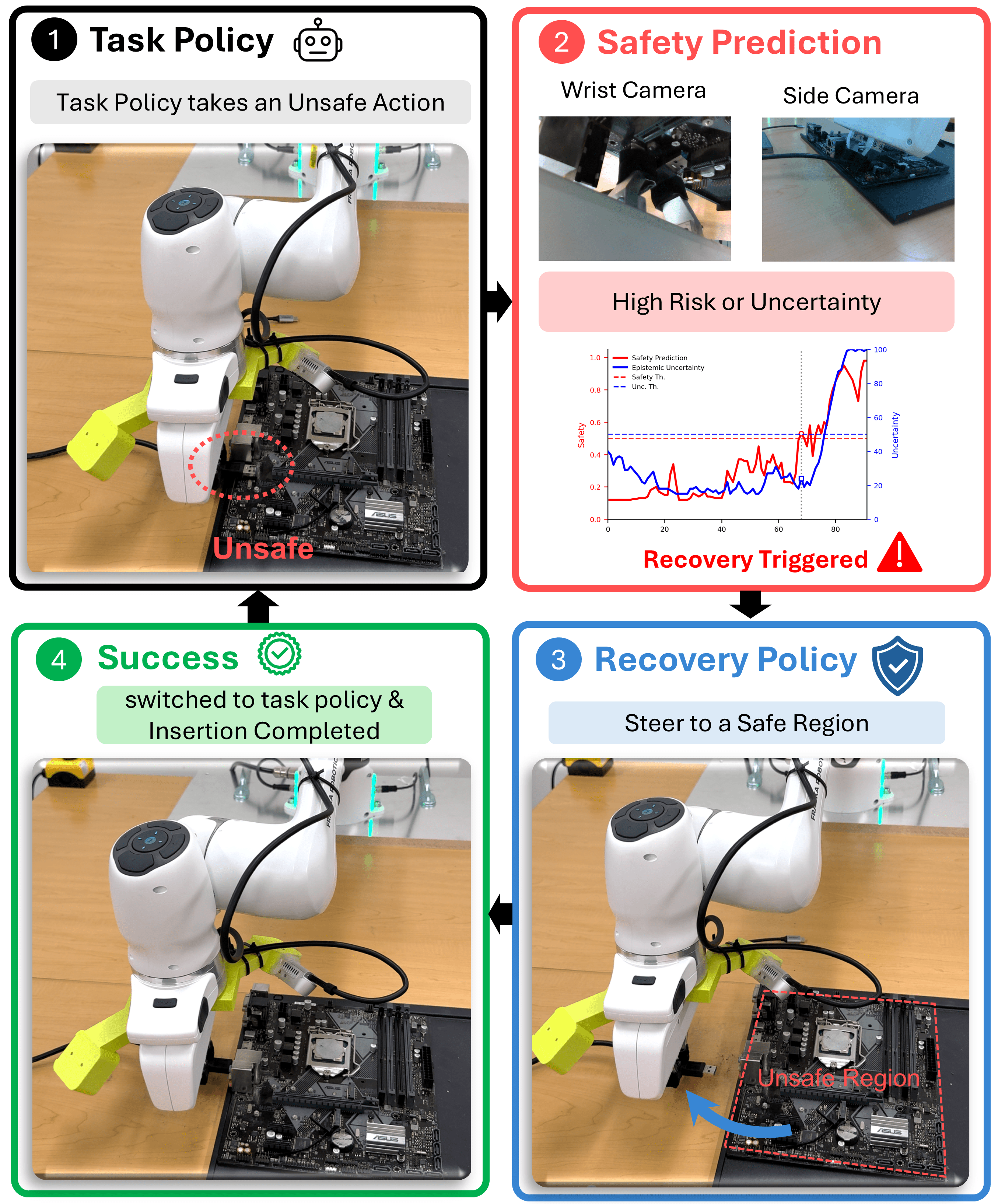}
    \vspace{-4 pt}
    \caption{\small 
Illustration of SEVI on the real-world USB insertion task. When the task policy attempts an unsafe insertion, SEVI detects elevated safety risk or epistemic uncertainty and triggers the recovery policy. The recovery policy steers the robot toward a safer region before returning control to the task policy for continued online RL training.
    }
    \vspace{-20pt}
    \label{fig:teaser}
\end{figure}

This challenge is particularly pronounced in complex tasks, where safety depends on perception, object geometry, contact dynamics, and the evolving task policy \cite{zhu2020ingredientsrealworldroboticreinforcement}. Existing real-world RL pipelines provide an effective foundation for online learning with prior data and off-policy updates. However, they do not explicitly address the autonomous intervention problem: \textit{determining when a proposed task action should be rejected and replaced by safe recovery behavior}. One way to address this intervention problem is to augment the task policy with a recovery mechanism. Recovery-based safe RL methods instantiate this idea by switching to a recovery policy when a learned safety critic predicts future constraint violations \cite{thananjeyan2021recovery}. A key limitation is that these intervention decisions rely on the reliability of the learned safety estimate, which may degrade under partial observations or in state-action regions poorly represented in prior data.

Existing approaches to uncertainty-aware safety attempt to detect unreliable predictions under distribution shift \cite{seo2025uncertainty}. World-model-based safety filters use learned latent dynamics to imagine future outcomes and identify potential out-of-distribution (OOD) failures \cite{Nakamura_2025}. Although this is a promising direction, it relies on accurate long-horizon prediction and sufficient coverage of safety-critical interactions. In complex manipulation, limited unsafe data and partial observability can cause learned dynamics to produce optimistic predictions in unfamiliar regions, leading unsafe actions to be classified as safe. Although sampling-based uncertainty methods can improve reliability estimation, ensembles\cite{lakshminarayanan2017simplescalablepredictiveuncertainty} and Monte Carlo dropout\cite{gal2016dropoutbayesianapproximationrepresenting} require multiple models or repeated forward passes, which increases latency for real-time robot control \cite{amini2020deep}.

To address this problem, we propose Safe Evidential Real-World Reinforcement Learning (SEVI), a framework for uncertainty-aware autonomous intervention during online real-world RL. SEVI introduces an Evidential Safety Critic (ESC), which models safety prediction as evidence accumulation and estimates both future constraint-violation risk and epistemic uncertainty in a single forward pass. Unlike sampling-based uncertainty methods, the ESC does not require ensembles or repeated stochastic inference, making it suitable for low-latency safety decisions. During online training, SEVI evaluates each action proposed by the task policy using both predicted risk and prediction reliability. If the action is predicted to be unsafe or the prediction is insufficiently supported by evidence, SEVI overrides the task action with an offline-trained recovery policy that guides the robot toward a lower-risk region as illustrated in Fig.~\ref{fig:teaser}. This intervention mechanism reduces unsafe interactions without modifying the underlying off-policy RL objective.

We evaluate SEVI on two simulated manipulation tasks and a real-world USB insertion task with a Franka robot. The experiments show that SEVI improves safety prediction under distribution shift, provides better-calibrated uncertainty than dropout and ensemble baselines, and reduces unsafe events during online real-world RL while maintaining strong task performance. In summary, this paper makes the following contributions:
\begin{itemize}
    \item 
    We introduce a modular real-world RL safety architecture that decouples task-policy learning from execution-time safety intervention while preserving the underlying task-learning objective.

    \item
    We propose an Evidential Safety Critic that estimates both future constraint-violation risk and epistemic uncertainty in a single forward pass, enabling low-latency reliability-aware safety assessment.

    \item 
    We integrate evidential safety prediction with an offline-trained recovery policy to enable autonomous intervention during online real-world RL with prior data, while remaining compatible with standard off-policy RL pipelines.
\end{itemize}

\section{Related Work}
\textbf{Real-world RL.}
Efficient learning from limited physical interaction remains challenging in the presence of high-dimensional observations, sparse rewards, and practical reset constraints. Recent progress has combined demonstrations or prior data with sample-efficient off-policy RL, enabling policies to be trained directly on physical robots within practical time budgets \cite{ball2023efficientonlinereinforcementlearning}. Sample Efficient Robotic Reinforcement Learning (SERL) provides a full-stack real-world RL pipeline that integrates robot control, reward inference, asynchronous training, and Reinforcement Learning with Prior Data (RLPD) off-policy updates for manipulation tasks \cite{luo2024serl}. These systems demonstrate that real-world RL can be practical when algorithmic and system-level components are carefully integrated. For complex manipulation tasks, online training can still require long hardware sessions with continuous human monitoring, resets, or corrective intervention. This supervision burden limits scalability, especially when unsafe behavior can occur during exploration. SEVI targets this bottleneck by adding an autonomous safety layer that rejects risky or unreliable task actions and invokes recovery during online training.

\textbf{Safe RL.}
Classical safe RL commonly formulates safety through constrained policy optimization~\cite{achiam2017constrained}, while control-theoretic approaches use control barrier functions or Hamilton--Jacobi reachability to enforce safety~\cite{ames2019control,chen2023hamilton}. Predictive and latent safety filters extend safety reasoning to learned models and high-dimensional observations~\cite{Nakamura_2025}. Such approaches often rely on predefined constraints, dynamics models, or explicit safety specifications. Recovery-based methods are more closely related to SEVI, as they override risky task actions before unsafe states are reached~\cite{thananjeyan2021recovery}. However, these methods typically trigger intervention from predicted risk alone, which may become unreliable under partial observations, policy-induced distribution shift, or limited data coverage. SEVI additionally considers the reliability of the safety estimate when determining whether recovery is required.

\textbf{Uncertainty and Failure Prediction.}
Uncertainty estimation is widely used to identify unreliable predictions in learned robotic systems. Monte Carlo dropout and deep ensembles estimate epistemic uncertainty through stochastic inference or model disagreement~\cite{gal2016dropoutbayesianapproximationrepresenting,lakshminarayanan2017simplescalablepredictiveuncertainty}. These methods are effective, but they require repeated forward passes or multiple models, which increases latency for real-time robot control. Recent failure-detection methods monitor deviations from successful behavior. For example, FAIL-Detect formulates runtime failure detection as sequential OOD detection and uses conformal prediction to detect failures from successful rollouts alone~\cite{xu2025can}. Such methods improve runtime awareness, but they mainly serve as monitors and do not directly provide autonomous intervention for online RL. Evidential learning provides a sampling-free alternative for uncertainty estimation. Instead of sampling over network weights or model instances, it models prediction as evidence accumulation and estimates uncertainty from the parameters of an evidential distribution~\cite{amini2020deep}. Building on this formulation, our Evidential Safety Critic predicts both future violation risk and epistemic uncertainty in a single forward pass. This design connects uncertainty estimation to action selection: recovery is triggered when an action is either high-risk or insufficiently supported by evidence.
\begin{figure*}[t]
    \centering
    \includegraphics[width=\textwidth]{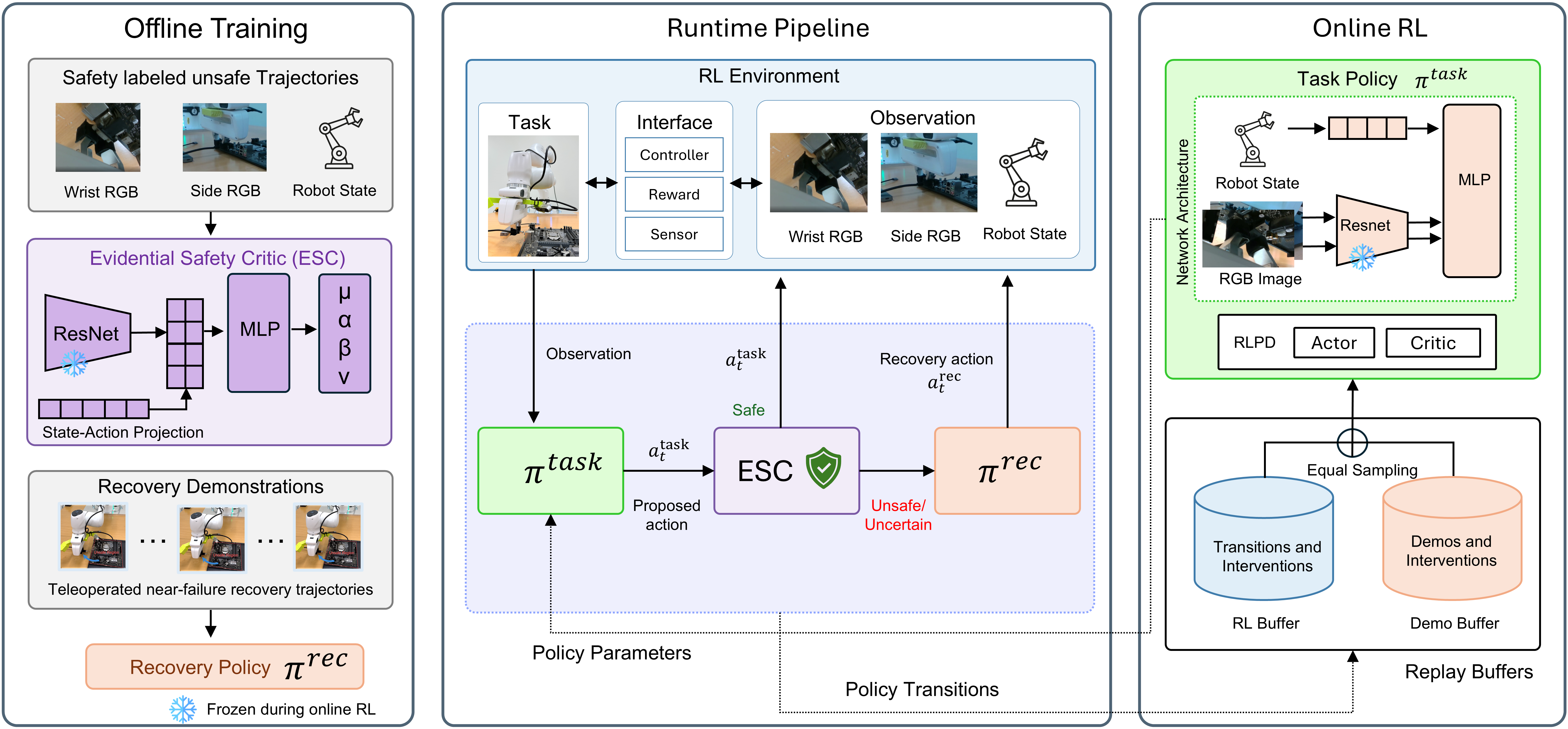} 
    \vspace{-2pt}
    \caption{\small
    \textbf{Overview of the SEVI framework}. In offline training, failure and recovery demonstrations are used to train the Evidential Safety Critic and recovery policy. During the runtime pipeline, the ESC evaluates task-policy actions using predicted risk and epistemic uncertainty and switches to the recovery policy when needed. In online RL, RLPD updates the task policy using executed transitions from the RL buffer together with prior demonstrations from the demo buffer.}
    \vspace{-8pt}
    \label{fig:Method Overview}
\end{figure*}

\section{Problem formulation}
We consider safe real-world RL for safety-critical robotic manipulation, where a robot learns a policy through physical interaction while minimizing unsafe events during training. We model the task-learning problem as a Markov Decision Process (MDP) $\mathcal{M} = (\mathcal{S}, \mathcal{A},\rho, \mathcal{P}, r, \gamma),$where $s \in \mathcal{S}$ is the state observation (e.g. image observation with the robot’s proprioceptive state information), $a \in \mathcal{A}$ denotes the control action (e.g. the desired end effector pose), $\rho(s_o)$ is the distribution over initial states, $\mathcal{P}$ is the unknown and potentially stochastic transition probabilities that depend on the system dynamics, $r: \mathcal{S} \times \mathcal{A} \rightarrow \mathbb{R}$ is the reward function which encodes the task and $ \gamma \in [0,1)$ is the discounted factor.
\begin{equation}
J_{\mathrm{task}}(\pi_{\mathrm{task}})
=
\mathbb{E}_{\pi_{\mathrm{task}}, P}
\left[
\sum_{t=0}^{\infty} \gamma^t r(s_t, a_t)
\right].
\end{equation}
Directly optimizing this objective on hardware can lead to unsafe robot-object interactions that require intervention. To model such events, we introduce a binary constraint signal $c_t \in \{0,1\}$, where $c_t = 1$ indicates a safety violation and $c_t = 0$ otherwise. In simulation, this signal can be obtained from privileged task information, such as object toppling, collision, or workspace violation. On hardware, it is obtained from task-specific safety observations, such as force thresholds, keep-out regions, or failure annotations.
Following prior safety-critic formulations \cite{thananjeyan2021recovery}, we define the risk value function as the discounted probability of future constraint violation after executing action $a_t$ in state $s_t$ and then following policy $\pi$:
\begin{equation}
Q_{\mathrm{risk}}^{\pi}(s_t,a_t)=\mathbb{E}_{\pi,P}\left[\sum_{k=0}^{\infty}\gamma_{\mathrm{risk}}^{k}c_{t+k}\mid s_t,a_t\right],
\label{eq:risk_value}
\end{equation}
where $\gamma_{\mathrm{risk}}\in[0,1)$ is the risk discount factor. This formulation assigns high risk not only to actions that are immediately unsafe, but also to actions that may lead to future constraint violations. The corresponding Bellman backup is
\begin{equation}
\begin{aligned}
Q_{\mathrm{risk}}^{\pi}(s_t,a_t)
&=
c_t
+
(1-c_t)\gamma_{\mathrm{risk}} \\
&\quad \times
\mathbb{E}_{a_{t+1}\sim\pi(\cdot\mid s_{t+1})}
\left[
Q_{\mathrm{risk}}^{\pi}(s_{t+1},a_{t+1})
\right].
\end{aligned}
\label{eq:risk_bellman}
\end{equation}
The factor $(1-c_t)$ treats violations as absorbing and propagates failure information backward along trajectories. In principle, safe policy learning can be viewed as maximizing task return while keeping predicted risk below a tolerance,
\begin{equation}
Q_{\mathrm{risk}}^{\pi}(s_t,a_t)\leq\epsilon_{\mathrm{risk}}.
\label{eq:risk_constraint}
\end{equation}
Although this risk constraint provides a natural formulation for safe RL, we do not optimize it through direct constrained policy optimization. Instead, we formulate safety as an execution-time filtering problem, where each action proposed by $\pi_{\mathrm{task}}$ is either executed or replaced by recovery behavior. The filtering decision depends on both the predicted future risk and the reliability of that prediction under limited data and distribution shift.

\section{Method}
We introduce Safe Evidential Real-World Reinforcement Learning (SEVI), a modular framework for uncertainty-aware safety intervention during online real-world RL. As shown in Fig.~\ref{fig:Method Overview}, SEVI combines three components: an Evidential Safety Critic for risk and uncertainty estimation, an offline-trained recovery policy for corrective intervention, and a task policy trained with sample-efficient off-policy RL using prior data. At runtime, each action proposed by the task policy is evaluated before execution. If the predicted risk or epistemic uncertainty exceeds a threshold, the proposed action is replaced by a recovery action. This design separates task learning from safety intervention while modifying only the executed action.

\subsection{Evidential Safety Critic (ESC)}
\label{sec:esc}

The ESC provides a reliability-aware safety estimate for execution-time intervention. It predicts a posterior predictive distribution over the temporal-difference risk target, from which we obtain both the expected future constraint-violation risk and epistemic uncertainty in a single forward pass. This enables the safety layer to reject actions that are either high-risk or poorly supported by evidence. 

For a transition $(s_t,a_t,s_{t+1},c_t)$, we define the risk Bellman target as
\begin{equation}
\begin{aligned}
    y_t
    =
    c_t
    +
    (1-c_t)\gamma_{\mathrm{risk}}
    \mathbb{E}_{a_{t+1}\sim \pi(\cdot|s_{t+1})}
    \left[
        \hat{Q}^{-}_{\mathrm{risk}}(s_{t+1},a_{t+1})
    \right],
\end{aligned}
\label{eq:esc-risk-target}
\end{equation}
where $c_t\in\{0,1\}$ is the safety-violation label and $\gamma_{\mathrm{risk}}\in[0,1)$ is the risk discount factor. If \(c_t=1\), the factor $(1-c_t)$ treats safety violations as absorbing, preventing the target from bootstrapping beyond a violation. The target prediction $\hat{Q}^{-}_{\mathrm{risk}}$ is computed using a slowly updated target critic with parameters $\bar{\phi}$, following standard temporal-difference learning. Since $y_t\in[0,1]$ is a bounded violation-probability target, we train the evidential critic in logit space. We first clip the Bellman target for numerical stability and define
\begin{equation}
    \bar{y}_t = \mathrm{clip}(y_t,\epsilon,1-\epsilon),
    \qquad
    z_t = \mathrm{logit}(\bar{y}_t).
    \label{eq:esc-logit-target}
\end{equation}
The transformed target $z_t$ is modeled as a noisy scalar observation generated from a Gaussian likelihood with unknown mean and variance,
\begin{equation}
    z_t \sim \mathcal{N}(\mu_t,\sigma_t^2).
    \label{eq:esc-gaussian-likelihood}
\end{equation}
Following deep evidential regression~\cite{amini2020deep}, we place a Normal--Inverse--Gamma (NIG) prior over the unknown likelihood parameters $(\mu_t,\sigma_t^2)$:
\begin{equation}
    \mu_t \sim \mathcal{N}(m_t,\sigma_t^2\nu_t^{-1}),
    \qquad
    \sigma_t^2 \sim \Gamma^{-1}(\alpha_t,\beta_t).
    \label{eq:esc-nig-prior}
\end{equation}
For each state--action pair, the critic network outputs the corresponding evidential parameters,
\begin{equation}
    (m_t,\nu_t,\alpha_t,\beta_t)
    =
    f_{\phi}(s_t,a_t),
    \label{eq:esc-nig-params}
\end{equation}
where $m_t$ is the predictive mean in logit space. The remaining three parameters define the evidential distribution over the unknown likelihood parameters. Specifically, $\nu_t$ controls the strength of evidence supporting the mean estimate $m_t$, $\alpha_t$ is the shape parameter of the inverse-gamma distribution over the variance, and $\beta_t$ is its scale parameter. The predicted future violation risk is obtained by mapping the expected latent mean back to probability space:
\begin{equation}
    \hat{Q}_{\mathrm{risk}}(s_t,a_t)
    =
    \sigma(m_t),
    \label{eq:esc-risk-pred}
\end{equation}
where $\sigma(\cdot)$ denotes the sigmoid function. The target critic uses the same mapping,
\begin{equation}
    \hat{Q}^{-}_{\mathrm{risk}}(s_t,a_t)
    =
    \sigma(m_{\bar{\phi}}(s_t,a_t)),
    \label{eq:esc-target-risk-pred}
\end{equation}
where $m_{\bar{\phi}}$ is the predictive mean produced by the target network. Marginalizing over the unknown likelihood parameters $(\mu_t,\sigma_t^2)$ gives the Student-$t$ posterior predictive distribution,
\begin{equation}
\begin{aligned}
    p(z_t \mid m_t,\nu_t,\alpha_t,\beta_t)
    =
    \mathrm{St}
    \left(
        z_t;
        m_t,
        \frac{\beta_t(1+\nu_t)}{\nu_t\alpha_t},
        2\alpha_t
    \right).
\end{aligned}
\label{eq:esc-student-t}
\end{equation}
Here, $m_t$ is the location parameter, $\frac{\beta_t(1+\nu_t)}{\nu_t\alpha_t}$ is the Student-$t$ scale parameter, and $2\alpha_t$ is the degrees of freedom. The negative log-likelihood of this predictive distribution defines the evidential regression loss,
\begin{equation}
    L_{\mathrm{NIG}}
    =
    -\log p(z_t \mid m_t,\nu_t,\alpha_t,\beta_t).
    \label{eq:esc-nig-nll}
\end{equation}
To discourage overconfident incorrect predictions, we use the evidence regularizer
\begin{equation}
    L_{\mathrm{reg}}
    =
    |z_t-m_t|\,(2\nu_t+\alpha_t).
    \label{eq:esc-evidence-regularizer}
\end{equation}
This term penalizes high evidence when the transformed target $z_t$ is far from the predictive mean $m_t$, discouraging overconfident risk estimates in poorly modeled state--action regions. The final evidential TD loss is
\begin{equation}
\begin{aligned}
    L_{\mathrm{ESC}}
    (s_t,a_t,s_{t+1},c_t;\phi)
    =
    L_{\mathrm{NIG}}
    +
    \lambda_{\mathrm{evi}} L_{\mathrm{reg}},
\end{aligned}
\label{eq:esc-edl-loss}
\end{equation}
where $\lambda_{\mathrm{evi}}$ controls the strength of the evidence regularizer. The critic parameters are optimized over replay data by minimizing
\begin{equation}
\begin{aligned}
    J^{\mathrm{ESC}}_{\mathrm{risk}}(\phi)
    =
    \mathbb{E}_{(s_t,a_t,s_{t+1},c_t)\sim\mathcal{D}}
    \left[
        L_{\mathrm{ESC}}
        (s_t,a_t,s_{t+1},c_t;\phi)
    \right].
\end{aligned}
\label{eq:esc-objective}
\end{equation}
The NIG distribution provides a closed-form estimate of epistemic uncertainty. Since epistemic uncertainty corresponds to uncertainty over the latent predictive mean $\mu_t$, we compute
\begin{equation}
    \sigma^2_{\mathrm{epi}}(s_t,a_t)
    =
    \mathrm{Var}[\mu_t]
    =
    \frac{\beta_t}
    {\nu_t(\alpha_t-1)}.
    \label{eq:esc-epistemic}
\end{equation}
A large value of $\sigma^2_{\mathrm{epi}}(s_t,a_t)$ indicates that the critic has limited evidence for the current state--action pair, which can occur in underexplored or out-of-distribution regions. Thus, $\hat{Q}_{\mathrm{risk}}(s_t,a_t)$ estimates the likelihood of future constraint violation, while $\sigma^2_{\mathrm{epi}}(s_t,a_t)$ measures the reliability of that estimate. Both quantities are passed to the recovery gate for uncertainty-aware action filtering.

\subsection{Recovery Policy}
\label{sec:recovery-policy}

The recovery policy $\pi_{\mathrm{rec}}$ provides a fallback behavior when the task policy proposes an action that is either high-risk or insufficiently supported by the safety critic. At each timestep, the task policy first proposes an action $a_t^{\mathrm{task}}\sim \pi_{\mathrm{task}}(\cdot \mid s_t)$. The ESC then evaluates the pair $(s_t,a_t^{\mathrm{task}})$ using the predicted risk $\hat{Q}_{\mathrm{risk}}(s_t,a_t^{\mathrm{task}})$ and epistemic uncertainty $\sigma^2_{\mathrm{epi}}(s_t,a_t^{\mathrm{task}})$. We define the recovery set as
\begin{equation}
\mathcal{T}_{\mathrm{rec}}
=
\left\{
(s,a)
\,\middle|\,
\hat{Q}_{\mathrm{risk}}(s,a)\ge \tau_{\mathrm{risk}}
\;\; \text{or} \;\;
\sigma^2_{\mathrm{epi}}(s,a)\ge \tau_{\mathrm{epi}}
\right\},
\label{eq:recovery-set}
\end{equation}
where $\tau_{\mathrm{risk}}$ and $\tau_{\mathrm{epi}}$ are safety and uncertainty thresholds calibrated on held-out offline data. If the proposed task action lies outside this set, it is executed. Otherwise, the action is replaced by the recovery policy:
\begin{equation}
a_t
=
\begin{cases}
a_t^{\mathrm{task}},
&
(s_t,a_t^{\mathrm{task}})\notin \mathcal{T}_{\mathrm{rec}},
\\[4pt]
a_t^{\mathrm{rec}},
&
(s_t,a_t^{\mathrm{task}})\in \mathcal{T}_{\mathrm{rec}},
\end{cases}
\label{eq:recovery-switch}
\end{equation}
where $a_t^{\mathrm{rec}}\sim \pi_{\mathrm{rec}}(\cdot \mid s_t)$. This rule leaves the task-policy objective unchanged and applies safety intervention only through execution-time action replacement. An ESC-only filter would require resampling or selecting another admissible task action, whereas $\pi_{\mathrm{rec}}$ provides an explicit corrective action that moves execution away from failure-prone regions before returning control to $\pi_{\mathrm{task}}$. Learning recovery behavior online is undesirable on physical robots because recovery actions are needed near failure boundaries, where exploratory behavior can cause damage or require human intervention. We therefore train $\pi_{\mathrm{rec}}$ offline using a dataset of human teleoperated recovery demonstrations $\mathcal{D}_{\mathrm{rec}}$, collected from near-failure states back toward low-risk operating regions. To align recovery with the learned safety model, we relabel recovery data using an ESC-shaped reward based on predicted risk:
\begin{equation}
r_t^{\mathrm{rec}}
=
-\hat{Q}_{\mathrm{risk}}(s_t,a_t).
\label{eq:recovery-reward}
\end{equation}
The recovery reward in Eq.~\eqref{eq:recovery-reward} depends only on the predicted mean risk and does not incorporate epistemic uncertainty. Epistemic uncertainty is used exclusively by the recovery gate to determine when recovery is activated, thereby separating the recovery objective from the uncertainty-aware intervention criterion. We train  $\pi_{\mathrm{rec}}$ with TD3+BC\cite{fujimoto2021minimalistapproachofflinereinforcement} on $\mathcal{D}_{\mathrm{rec}}$. The value-based objective encourages low-risk recovery behavior, while behavior cloning regularization keeps the learned policy close to demonstrated corrective actions and reduces extrapolation outside the recovery dataset. During online real-world RL, the recovery policy acts as a stabilizing fallback. It is invoked only when the proposed task action enters the recovery set, and control returns to the task policy once the proposed task actions satisfy the risk and uncertainty thresholds.
\vspace{-1em}
\subsection{Integration with Online Real-World RL}
\label{sec:real-world-safe-rl}

In our implementation, the task policy $\pi_{\mathrm{task}}$ is trained with RLPD within the SERL framework \cite{luo2024serl}. The task learner uses prior demonstrations, off-policy actor-critic updates, and online replay in the standard SERL pipeline. SEVI leaves the task reward, actor objective, critic update, and replay sampling procedure unchanged. 
At each timestep, the task policy proposes an action $a_t^{\mathrm{task}}\sim \pi_{\mathrm{task}}(\cdot\mid s_t)$. 
The executed transition is stored in the online replay buffer and reused for task-policy updates. Thus, the task policy continues to optimize the original task objective, while the collected data are biased toward lower-risk interactions.
SEVI is intended as a practical execution-time safety layer, similar in spirit to Recovery RL and runtime shielding. Since both the safety critic and recovery policy are learned from finite data, SEVI does not provide formal guarantees of zero violations. Instead, its goal is to reduce unsafe interactions during online real-world RL while preserving the sample efficiency and task-learning behavior of the underlying RL algorithm.
\begin{figure}[t]
    \centering
    \includegraphics[width=\columnwidth]{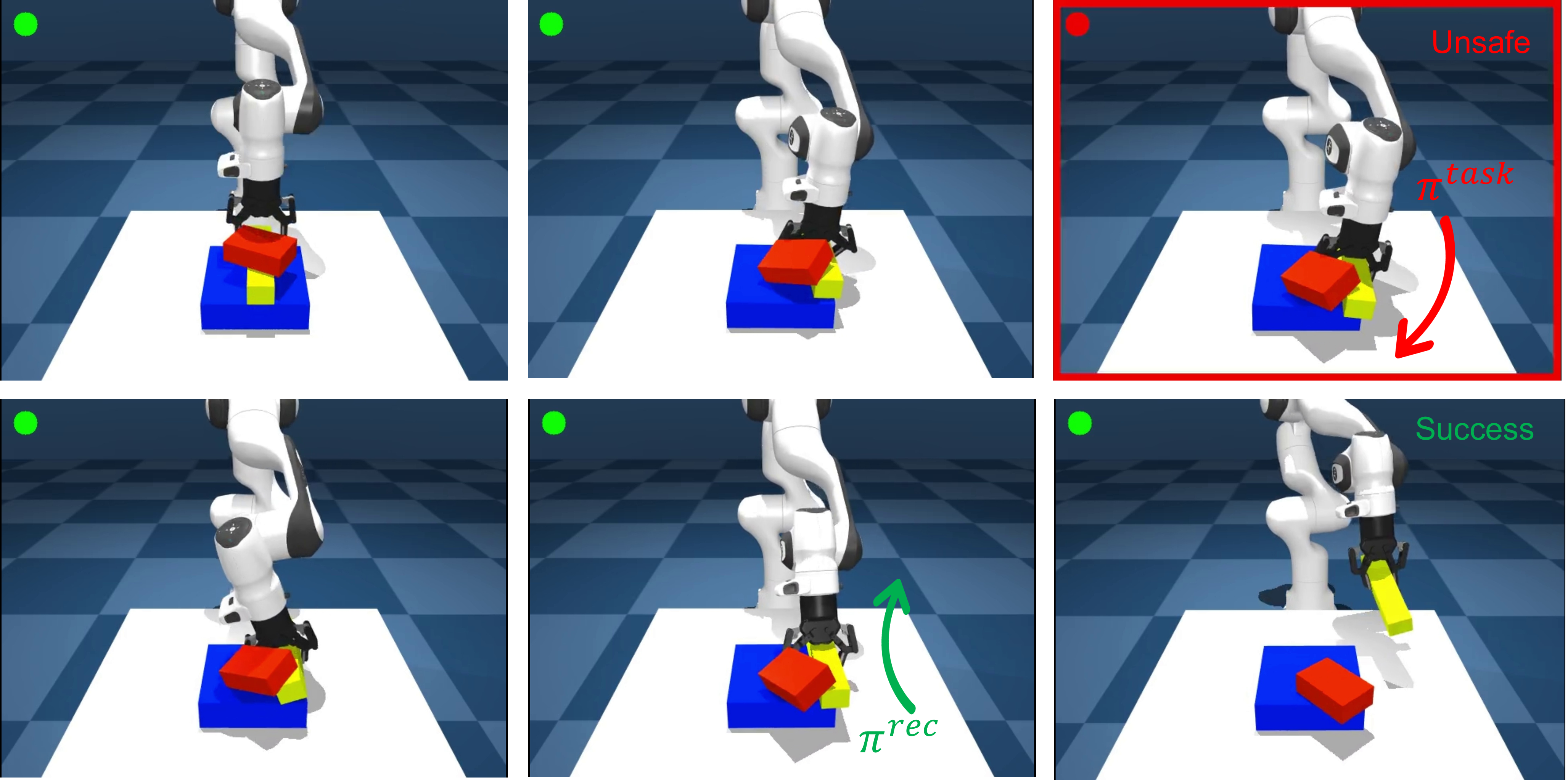} 
    \caption{\small Block Plucking example of SEVI task–recovery switching. An unsafe $\pi^{\mathrm{task}}$ action that could destabilize the stack (red) triggers $\pi^{\mathrm{rec}}$, which returns the robot to a safer configuration before task execution resumes successfully (green).}
    \vspace{-18pt}
    \label{fig:simulation_task}
\end{figure}

\section{Results}
\label{sec:results}

Our experiments are designed to assess three critical aspects of SEVI:
\textbf{(Q1)} Can ESC provide reliable safety prediction under distribution shift with real-time inference?
\textbf{(Q2)} Does epistemic uncertainty improve recovery gating beyond predicted risk alone?
\textbf{(Q3)} Can SEVI reduce unsafe interactions during online real-world RL while preserving task performance?

We first evaluate safety prediction under in-distribution (ID) and out-of-distribution (OOD) conditions in both simulation and hardware to address \textbf{(Q1)} in Sec.~\ref{sec:q1}. We then examine \textbf{(Q2)} in Sec.~\ref{sec:q2} by analyzing both the quality of epistemic uncertainty and its effect on risk-based recovery gating. Finally, we address \textbf{(Q3)} in Sec.~\ref{sec:q3} through closed-loop online RL experiments in simulation and on a real robot, evaluating whether SEVI improves safety without compromising task success.
\vspace{-10pt}
\subsection{Evaluation Setup}

\textbf{Baselines.}
We evaluate SEVI with two groups of baselines. The first group measures closed-loop safe task learning and includes SERL~\cite{luo2024serl}, SAC~\cite{haarnoja2019softactorcriticalgorithmsapplications}, SAC-C~\cite{achiam2017constrained}, SQRL-style recovery~\cite{srinivasan2020learningsafedeeprl,thananjeyan2021recovery}, SEVI without uncertainty, and full SEVI. SERL and SAC evaluate task learning without execution-time recovery filtering, while SAC-C evaluates constraint-penalized policy learning. The SQRL-style recovery baseline uses a point-estimate safety critic trained with the SQRL future-violation risk objective, but uses the same offline-trained recovery policy as SEVI. This keeps the recovery behavior fixed and isolates the effect of the safety critic used for intervention. SEVI without uncertainty uses the evidential critic but triggers recovery only from predicted mean risk, while full SEVI triggers recovery using both predicted risk and epistemic uncertainty. The second group evaluates safety prediction and uncertainty estimation independently of closed-loop task performance. We compare the ESC against a point-estimate SQRL-style critic~\cite{srinivasan2020learningsafedeeprl}, MC Dropout~\cite{gal2016dropoutbayesianapproximationrepresenting}, and Deep Ensembles~\cite{lakshminarayanan2017simplescalablepredictiveuncertainty}. These baselines use comparable observation-action inputs and safety-label supervision, but differ in how prediction reliability is represented. This allows us to separately evaluate the closed-loop benefit of uncertainty-aware recovery and the standalone quality of the proposed safety critic.

\textbf{Platforms and Environments.}
We evaluate SEVI on two visual manipulation tasks in MuJoCo and one hardware task using a Franka Research 3 arm. In the \textbf{Block Plucking} task, the robot must remove the middle block from a three-block stack while keeping the top block supported by the bottom block as shown in Fig.~\ref{fig:simulation_task}. The policy observes two $128\times128$ RGB views from front and wrist cameras together with a $16$D state consisting of the 3D TCP position, 3D TCP linear velocity, gripper open/close state, and the 3D Cartesian positions of the three cubes; each episode runs at $50$~Hz for a $300$-step horizon. In \textbf{Transport Nut}, the robot must transport a square nut and insert it onto a target peg. Safety violations are defined as intervention-requiring events, including dropping the nut, moving it outside the reachable workspace, or causing insertion jamming due to large pose misalignment. For hardware evaluation, the \textbf{USB Insertion} task requires the robot to pick up a USB connector and insert it into a target slot while avoiding keep-out regions on the motherboard and excessive contact forces. The hardware policy observes dual $128\times128$ RGB views with a $19$D state consisting of the 6D TCP pose (position and orientation), 6D TCP velocity (linear and angular), gripper open/close state, 3D TCP force, and 3D TCP torque, and executes at $10$~Hz for a $120$-step horizon. Distribution shift is evaluated through visual and task-configuration perturbations in simulation, and through distractor objects and cable-layout variations in the USB Task. All simulation training and real-robot experiments are run on a workstation with an Intel Core i9-14900KF CPU, 64~GB RAM, and an NVIDIA RTX 4090 GPU with 24~GB memory, running Ubuntu 22.04.5 LTS.

\textbf{Evaluation Metrics and Protocol.}
In simulation, each method is trained with 5 independent random seeds, and the final policy from each seed is evaluated over 20 rollouts, yielding 100 evaluation rollouts per method for each task. For hardware evaluation, each method is trained in three independent 3-hour real-world online RL sessions on the USB Insertion task, corresponding to 9 hours of hardware training per method. For each run, the final policy is evaluated over 50 hardware rollouts, while unsafe events are counted over the entire online training session. Following real-world RL evaluation practice, we report the success rate of the final policy over 50 evaluation rollouts, while unsafe events and recovery interventions are counted over the entire online training session. This separates final task performance from the safety cost incurred during real-world policy improvement. We report three groups of metrics. Following prior runtime failure-prediction protocols~\cite{xu2025can,romer2025fiper}, we report trajectory-level TPR, TNR, balanced accuracy (BA), normalized detection time (DT), and inference latency. BA is computed for each seed as $\mathrm{BA}=\frac{1}{2}(\mathrm{TPR}+\mathrm{TNR})$, and then reported as the mean and standard deviation across seeds. For uncertainty quality, we report confidence AUROC/APR, OOD AUROC/APR, expected calibration error (ECE), and Brier score~\cite{itkina2022interpretableselfawareneuralnetworks}. For closed-loop execution, we report task success rate, total unsafe events during training, recovery intervention rate, and cycle time. Closed-loop simulation and hardware results are reported as mean and standard deviation across five and three independent runs, respectively.

\textbf{Implementation Details.} The implementation details of the proposed modules are as follows:
\newline \textbf{Evidential Safety Critic (ESC):} The ESC uses a twin-critic architecture to reduce overconfident risk estimates. Dual-camera RGB observations from wrist and front views are processed by a frozen ResNet-v1-10 encoder with shared weights. The visual embeddings are concatenated with a linear projection of the proprioceptive state and action. The fused representation is passed through a two-layer MLP with 256 hidden units, Swish activations, and LayerNorm, followed by an evidential output head that predicts the NIG parameters $(m,\nu,\alpha,\beta)$. For conservative target computation and action filtering, we take the maximum predicted risk and maximum epistemic uncertainty across the two critics. The thresholds $\tau_{\mathrm{risk}}$ and $\tau_{\mathrm{epi}}$ are calibrated once before online RL using 50 successful and 50 failure validation rollouts and kept fixed thereafter. The calibrated pairs are Block Plucking $(0.44, 0.56)$, Transport Nut $(0.61, 0.64)$, and USB Insertion $(0.65, 0.45)$. Perturbing both thresholds by $\pm 20\%$ changes the success rate by less than $8\%$, $9\%$, and $5\%$, respectively. Safety labels $c_t$ are obtained from privileged simulation states or hardware failure annotations.

\textbf{Recovery Policy ($\pi_{\mathrm{rec}}$):} The recovery policy is trained offline with TD3+BC on teleoperated recovery trajectories collected near task-specific failure boundaries. We use 150 recovery trajectories for each simulation task and 200 trajectories for the hardware task. The actor and critic use two-layer MLPs with 256 hidden units, learning rate $1\times10^{-4}$, discount factor $\gamma=0.95$, and behavior cloning weight $\lambda_{\mathrm{BC}}=2.5$. The same recovery policy is used across recovery-based variants to isolate the effect of the safety critic and gating rule.

\textbf{Task Policy ($\pi_{\mathrm{task}}$):} The task policy is trained with the SERL off-policy RL pipeline using 50 successful demonstrations as prior data. The actor and critic use two-layer MLPs with 256 hidden units, learning rate $3\times10^{-4}$, a 200k-transition replay buffer, update-to-data ratio of 4, and automatic entropy tuning with target entropy $-\dim(\mathcal{A})/2$.

\begin{figure}[t]
    \centering

    \begin{minipage}[c]{0.43\columnwidth}
        \centering
        \includegraphics[width=\linewidth]
        {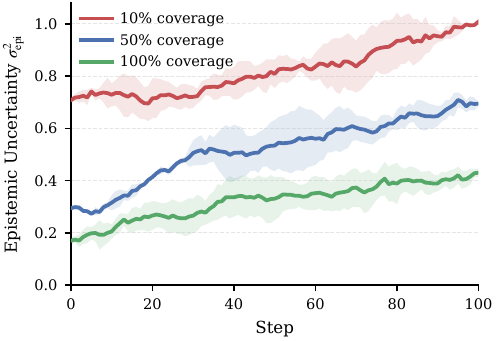}
        
        \vspace{-2pt}
        {\small (a)}
    \end{minipage}
    \hfill
    \begin{minipage}[c]{0.55\columnwidth}
        \centering
        \includegraphics[width=\linewidth]
        {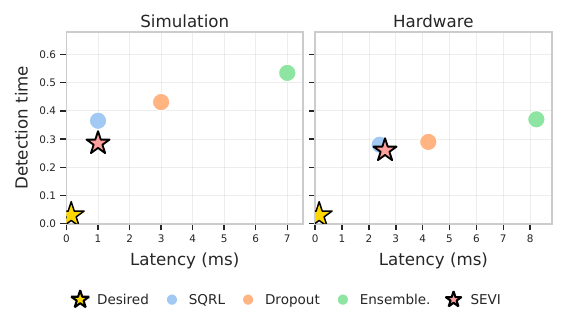}
        
        \vspace{-2pt}
        {\small (b)}
    \end{minipage}

    \vspace{-3pt}

    \caption{\small
    Characterization of the Evidential Safety Critic.
    {(a) Epistemic uncertainty under increasing training coverage in a fixed USB failure region; shading denotes $\pm1$ standard deviation across three ESC trainings.}
    (b) Detection efficiency of safety critics in simulation and hardware.
    }
    \label{fig:esc_characterization}
    \vspace{-8pt}
\end{figure}

\begin{table}[t]
\centering
\caption{\small
Trajectory-level safety prediction under ID and OOD conditions. SEVI achieves the highest balanced accuracy across tasks with consistent TPR--TNR balance.
}
\label{tab:safety_prediction}
\renewcommand{\arraystretch}{1.08}
\setlength{\tabcolsep}{3.2pt}
\scriptsize
\resizebox{\columnwidth}{!}{
\begin{tabular}{llccccc}
\toprule
Task & Metric & SQRL & Dropout & Ensemble & SEVI(w/o Unc.) & \textbf{SEVI} \\
\midrule

\multirow{3}{*}{Block ID}
& BA$\uparrow$  
& $.85{\pm}.01$ & $.82{\pm}.03$ & $.90{\pm}.03$ & $.83{\pm}.04$ & $\mathbf{.93{\pm}.03}$ \\
& TPR$\uparrow$ 
& $.77{\pm}.03$ & $.74{\pm}.04$ & $.86{\pm}.04$ & $.80{\pm}.05$ & $\mathbf{.92{\pm}.03}$ \\
& TNR$\uparrow$ 
& $.93{\pm}.03$ & $.90{\pm}.03$ & $\mathbf{.95{\pm}.02}$ & $.86{\pm}.04$ & $.94{\pm}.04$ \\

\midrule
\multirow{3}{*}{Block OOD}
& BA$\uparrow$  
& $.66{\pm}.03$ & $.69{\pm}.01$ & $.74{\pm}.05$ & $.66{\pm}.03$ & $\mathbf{.80{\pm}.02}$ \\
& TPR$\uparrow$ 
& $.61{\pm}.03$ & $.54{\pm}.03$ & $.63{\pm}.07$ & $.61{\pm}.04$ & $\mathbf{.73{\pm}.03}$ \\
& TNR$\uparrow$ 
& $.71{\pm}.03$ & $.84{\pm}.03$ & $.85{\pm}.04$ & $.72{\pm}.03$ & $\mathbf{.88{\pm}.03}$ \\

\midrule
\multirow{3}{*}{Nut ID}
& BA$\uparrow$  
& $.52{\pm}.02$ & $.59{\pm}.04$ & $.64{\pm}.04$ & $.52{\pm}.03$ & $\mathbf{.75{\pm}.03}$ \\
& TPR$\uparrow$ 
& $.23{\pm}.02$ & $.31{\pm}.06$ & $.39{\pm}.05$ & $.27{\pm}.03$ & $\mathbf{.66{\pm}.04}$ \\
& TNR$\uparrow$ 
& $.81{\pm}.03$ & $.86{\pm}.04$ & $\mathbf{.89{\pm}.05}$ & $.77{\pm}.05$ & $.83{\pm}.04$ \\

\midrule
\multirow{3}{*}{Nut OOD}
& BA$\uparrow$  
& $.49{\pm}.03$ & $.54{\pm}.03$ & $.59{\pm}.03$ & $.46{\pm}.02$ & $\mathbf{.65{\pm}.01}$ \\
& TPR$\uparrow$ 
& $.38{\pm}.02$ & $\mathbf{.55{\pm}.04}$ & $.34{\pm}.02$ & $.38{\pm}.03$ & $.48{\pm}.03$ \\
& TNR$\uparrow$ 
& $.59{\pm}.05$ & $.54{\pm}.04$ & $\mathbf{.85{\pm}.05}$ & $.54{\pm}.04$ & $.83{\pm}.03$ \\

\midrule
\multirow{3}{*}{USB ID}
& BA$\uparrow$  
& $.70{\pm}.02$ & $.73{\pm}.01$ & $.79{\pm}.05$ & $.66{\pm}.04$ & $\mathbf{.84{\pm}.03}$ \\
& TPR$\uparrow$ 
& $.58{\pm}.04$ & $.67{\pm}.02$ & $.82{\pm}.07$ & $.52{\pm}.05$ & $\mathbf{.82{\pm}.04}$ \\
& TNR$\uparrow$ 
& $.81{\pm}.07$ & $.79{\pm}.02$ & $.75{\pm}.04$ & $.80{\pm}.05$ & $\mathbf{.86{\pm}.04}$ \\

\midrule
\multirow{3}{*}{USB OOD}
& BA$\uparrow$  
& $.58{\pm}.03$ & $.58{\pm}.05$ & $.64{\pm}.03$ & $.59{\pm}.03$ & $\mathbf{.74{\pm}.02}$ \\
& TPR$\uparrow$ 
& $.51{\pm}.06$ & $.40{\pm}.05$ & $.43{\pm}.05$ & $.54{\pm}.05$ & $\mathbf{.57{\pm}.02}$ \\
& TNR$\uparrow$ 
& $.65{\pm}.03$ & $.77{\pm}.05$ & $.85{\pm}.02$ & $.64{\pm}.04$ & $\mathbf{.91{\pm}.03}$ \\

\bottomrule
\end{tabular}
}
\vspace{-2mm}
\end{table}

\subsection{Safety Prediction under Distribution Shift}
\label{sec:q1}
We first evaluate the safety critic independently of closed-loop task learning. Table~\ref{tab:safety_prediction} reports trajectory-level safety prediction results across Block Plucking, Transport Nut, and USB Insertion under both in-distribution (ID) and out-of-distribution (OOD) conditions. SEVI achieves the highest balanced accuracy in all evaluated settings, improving over the point-estimate SQRL and MC Dropout, while remaining comparable to or better than Deep Ensembles. This is important because ensembles are a strong uncertainty baseline, but require multiple independently trained critics, whereas SEVI obtains both risk and epistemic uncertainty in a single forward pass. The gains are most pronounced under distribution shift. On Block Plucking OOD, SEVI improves balanced accuracy from $0.66$ for the SQRL-style critic to $0.80$, and on USB Insertion OOD from $0.58$ to $0.74$. Transport Nut remains the most difficult setting because failures can occur across grasping, transport, and insertion phases. Nevertheless, SEVI achieves the highest OOD balanced accuracy. The TPR and TNR values show that this improvement is not caused by uniformly conservative predictions. SEVI improves unsafe-trajectory detection while maintaining strong safe-trajectory rejection, which is necessary for online RL where excessive false positives can suppress useful exploration.

\begin{table}[t]
\centering
\caption{Detection time (DT) and inference latency of the evaluated safety critics corresponding to Fig.~\ref{fig:esc_characterization}(b).}
\label{tab:detection_latency}

\footnotesize
\renewcommand{\arraystretch}{1.25}
\setlength{\tabcolsep}{3.5pt}

\begin{tabular}{@{}lcccc@{}}
\toprule
& \multicolumn{2}{c}{Simulation (Block Plucking)}
& \multicolumn{2}{c}{Hardware (USB Insert)} \\
\cmidrule(lr){2-3}\cmidrule(lr){4-5}

Method
& DT$\downarrow$
& Lat. (ms)$\downarrow$
& DT$\downarrow$
& Lat. (ms)$\downarrow$ \\
\midrule

SQRL
& $0.37{\pm}0.03$
& $1.00{\pm}0.08$
& $0.28{\pm}0.02$
& $2.40{\pm}0.15$ \\

Dropout
& $0.43{\pm}0.04$
& $3.00{\pm}0.20$
& $0.29{\pm}0.03$
& $4.21{\pm}0.25$ \\

Ensemble
& $0.53{\pm}0.05$
& $7.00{\pm}0.40$
& $0.37{\pm}0.04$
& $8.23{\pm}0.50$ \\

\textbf{SEVI}
& $\mathbf{0.29{\pm}0.02}$
& $\mathbf{1.00{\pm}0.05}$
& $\mathbf{0.26{\pm}0.02}$
& $\mathbf{2.60{\pm}0.10}$ \\

\bottomrule
\end{tabular}
\end{table}




Fig.~\ref{fig:esc_characterization}(b) and Table~\ref{tab:detection_latency} show that SEVI achieves the lowest normalized detection time on both Block Plucking and USB Insertion while maintaining latency comparable to the point-estimate SQRL critic. Unlike SQRL, however, SEVI also provides epistemic uncertainty in the same forward pass; MC Dropout and Deep Ensembles obtain uncertainty estimates at substantially higher inference cost. Together with Table~\ref{tab:safety_prediction}, these results show that SEVI provides accurate safety prediction, early detection, and low-latency uncertainty estimation, addressing Q1.

\begin{table}[t]
\centering
\caption{\small Uncertainty estimation quality on the hardware task under ID and OOD conditions. SEVI provides stronger discrimination and calibration than baselines.}
\label{tab:uncertainty_quality}
\renewcommand{\arraystretch}{1.10}
\setlength{\tabcolsep}{4.8pt}
\scriptsize
\begin{tabular}{llccc}
\toprule
Set. & Metric & Dropout & Ensemble & \textbf{SEVI}\\
\midrule
\multirow{4}{*}{ID}
& AUROC $\uparrow$  
& $0.68$ 
& $0.88$ 
& $\mathbf{0.92}$ \\

& APR $\uparrow$    
& $0.74$ 
& $0.87$ 
& $\mathbf{0.96}$ \\

& ECE $\downarrow$  
& $0.29$ 
& $0.38$ 
& $\mathbf{0.07}$ \\

& Brier $\downarrow$
& $0.29$ 
& $0.28$ 
& $\mathbf{0.04}$ \\

\midrule
\multirow{4}{*}{OOD}
& AUROC $\uparrow$  
& $0.62$ 
& $0.89$ 
& $\mathbf{0.94}$ \\

& APR $\uparrow$    
& $0.25$ 
& $0.21$ 
& $\mathbf{0.39}$ \\

& ECE $\downarrow$  
& $0.13$ 
& $0.30$ 
& $\mathbf{0.09}$ \\

& Brier $\downarrow$
& $0.12$ 
& $0.19$ 
& $\mathbf{0.07}$ \\

\bottomrule
\end{tabular}
\vspace{-2mm}
\end{table}

\begin{figure}[t]
{
    \centering
    \includegraphics[width=\columnwidth]{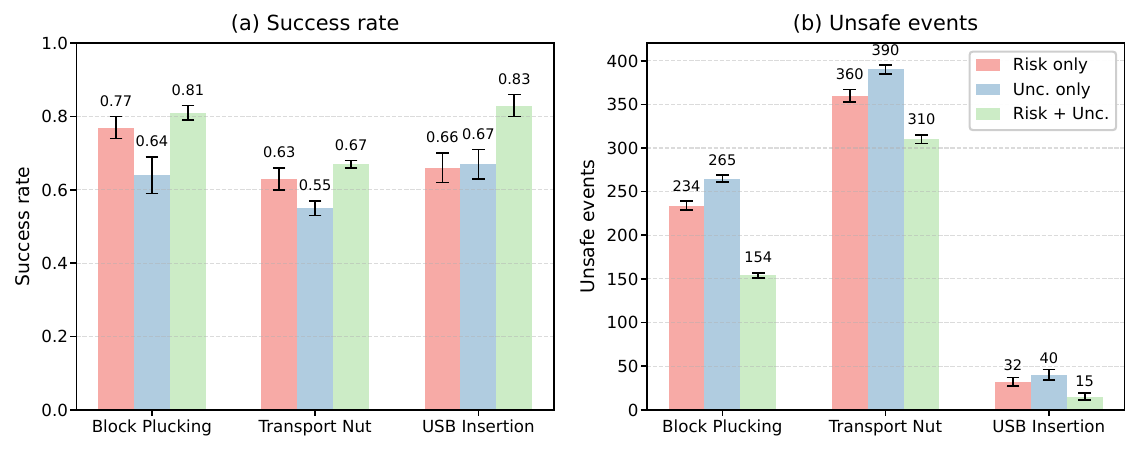} \vspace{-15pt}
    \caption{\small
    Effect of recovery-gating criteria on closed-loop execution. Combining risk and uncertainty gives the best safety--performance trade-off across tasks. \vspace{-15pt}
    }
    \label{fig:gating_ablation}
}
\end{figure}

\subsection{Uncertainty-Aware Recovery Gating}
\label{sec:q2}
To answer Q2, we evaluate whether epistemic uncertainty is well calibrated, sensitive to training-data coverage, and useful for downstream recovery decisions. Table~\ref{tab:uncertainty_quality} reports uncertainty estimation results on the USB insertion task. SEVI achieves the strongest performance under both ID and OOD settings, obtaining higher AUROC/APR and lower ECE/Brier score than dropout and ensemble baselines. These results indicate that the evidential critic produces uncertainty estimates that are both discriminative and well calibrated under distribution shift. 
To assess whether epistemic uncertainty reflects limited data support, we train the ESC using $10\%$, $50\%$, and $100\%$ of 200 trajectories collected from a fixed USB failure region and evaluate all models on the same 50 held-out trajectories. As shown in Fig.~\ref{fig:esc_characterization}(a), epistemic uncertainty decreases consistently as training coverage increases, providing empirical evidence that the estimate captures limited support in the training data.
Fig.~\ref{fig:gating_ablation} evaluates whether this uncertainty improves downstream action filtering. We compare three SEVI gating variants: risk-only, uncertainty-only, and risk-plus-uncertainty. The combined gate achieves the best safety--performance trade-off across all tasks. Relative to risk-only gating, it improves success by $4$ percentage points on Block Plucking and $17$ percentage points on USB Insertion, while reducing unsafe events by $34.2\%$ and $53.1\%$, respectively. On Transport Nut, the combined gate reduces unsafe events by $13.9\%$. Aggregated across tasks, risk-plus-uncertainty gating reduces unsafe events by $\mathbf{23.5\%}$ compared with risk-only gating and by $\mathbf{31.1\%}$ compared with uncertainty-only gating. To assess whether this benefit depends on the recovery behavior, Table~\ref{tab:effect_recovery_behavior} evaluates two qualitatively distinct USB recovery responses: Hit-and-Recover and Stuck-and-Recover. Risk-plus-uncertainty gating consistently improves task success and reduces unsafe events under both behaviors, indicating that the gating advantage persists across distinct recovery responses.
These results demonstrate that predicted risk and epistemic uncertainty provide complementary intervention signals. Predicted risk captures the likelihood of future constraint violations, whereas epistemic uncertainty identifies state--action regions in which the safety estimate is poorly supported by data. Their combination therefore enables more selective recovery intervention while preserving task performance.
\begin{figure}[t]
    \centering
    \includegraphics[width=\columnwidth]{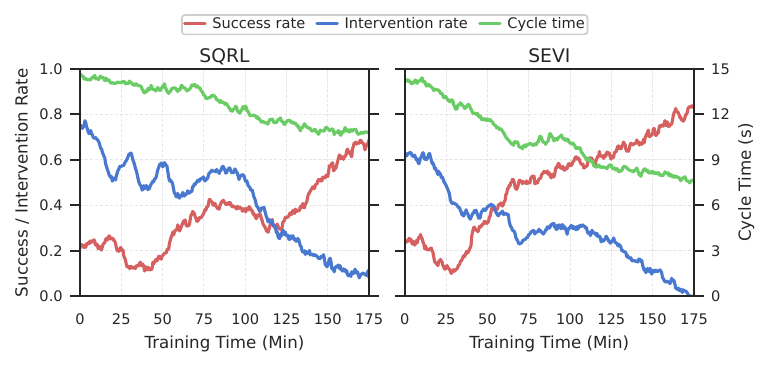}
    \vspace{-15pt}
    \caption{\small
    Learning curves from a single representative hardware training run on the USB insertion task, showing success rate, intervention rate, and cycle time for SQRL and SEVI during online training. SEVI reaches higher success while progressively reducing intervention rate and cycle time.}
    \label{fig:learning curve}
    \vspace{-2pt}
\end{figure}
\subsection{Safe Online Policy Learning}
\label{sec:q3}
To address Q3, we evaluate whether SEVI reduces unsafe interactions during online RL without slowing policy learning or overly restricting execution. Fig.~\ref{fig:learning curve} shows that SEVI achieves this balance more effectively than SQRL under the fixed interaction budget on USB insertion task. After $175$ minutes, SEVI reaches $0.84$ success with a $7.6$~s cycle time, compared with $0.69$ success and $11.4$~s for SQRL. {Repeated task--recovery switching can occur during early training when the task policy repeatedly proposes inadmissible actions; however, the declining intervention rate in Fig.~\ref{fig:learning curve} indicates that such switching becomes less frequent as the policy improves.} In contrast, SQRL exhibits larger intervention fluctuations and slower success improvement, suggesting that mean-risk filtering alone provides a weaker intervention signal.

\begin{table}[t]
\centering
\caption{Effect of recovery behavior on USB Insertion task.}
\label{tab:effect_recovery_behavior}

\footnotesize
\setlength{\tabcolsep}{4pt}
\renewcommand{\arraystretch}{1.30}

\begin{tabular}{@{}llcc@{}}
\toprule
Recovery Behavior & Gating & Success$\uparrow$ & Unsafe$\downarrow$ \\
\midrule

\multirow{2}{*}{Hit-and-Recover}
& Risk only & $0.64\pm0.03$ & $32\pm4$ \\
& Risk+Unc. & $\mathbf{0.82\pm0.03}$ & $\mathbf{16\pm4}$ \\

\midrule

\multirow{2}{*}{Stuck-and-Recover}
& Risk only & $0.64\pm0.01$ & $33\pm4$ \\
& Risk+Unc. & $\mathbf{0.83\pm0.03}$ & $\mathbf{15\pm5}$ \\

\bottomrule
\end{tabular}
\end{table}



Table~\ref{tab:task_performance} reports closed-loop task performance across simulation and hardware. {For USB Insertion, results are reported as mean and standard deviation across three independent hardware training runs}. SEVI provides the strongest safety--performance trade-off, achieving the highest success rate and the lowest unsafe-event count on all three tasks. Relative to SQRL, SEVI reduces unsafe events by $30.0\%$ on Block Plucking, $13.4\%$ on Transport Nut, and {$57.1\%$} on USB Insertion. SERL and SAC achieve reasonable task performance but accumulate substantially more unsafe events due to the absence of execution-time recovery filtering. SAC-C reduces violations through constraint penalties, but remains less effective than SEVI, indicating that fixed penalties are insufficient for policy-induced safety risks in complex manipulation tasks. Overall, these results support Q3: reliability-aware recovery reduces unsafe interactions during online policy learning while preserving task improvement and execution efficiency.

\begin{table}[t]
\centering
\caption{\small
Closed-loop task execution performance. Success is evaluated using the final policy, while unsafe events are counted throughout training.
}
\label{tab:task_performance}
\renewcommand{\arraystretch}{1.10}
\setlength{\tabcolsep}{3.2pt}
\scriptsize
\resizebox{\columnwidth}{!}{
\begin{tabular}{lcccccc}
\toprule
\multirow{2}{*}{Method}
& \multicolumn{2}{c}{Block Plucking}
& \multicolumn{2}{c}{Transport Nut}
& \multicolumn{2}{c}{USB Insertion} \\
\cmidrule(lr){2-3} \cmidrule(lr){4-5} \cmidrule(lr){6-7}
& Succ. $\uparrow$
& Unsafe $\downarrow$
& Succ. $\uparrow$
& Unsafe $\downarrow$
& Succ. $\uparrow$
& Unsafe $\downarrow$ \\
\midrule

SERL~\cite{luo2024serl}
& $0.67{\pm}.02$ & $270{\pm}7$
& $0.52{\pm}.01$ & $450{\pm}11$
& {$0.71{\pm}.03$} & {$56{\pm}7$} \\

SAC~\cite{haarnoja2019softactorcriticalgorithmsapplications}
& $0.63{\pm}.01$ & $298{\pm}5$
& $0.49{\pm}.01$ & $467{\pm}9$
& {$0.61{\pm}.03$} & {$49{\pm}4$} \\

SAC-C~\cite{achiam2017constrained}
& $0.71{\pm}.03$ & $281{\pm}3$
& $0.55{\pm}.02$ & $410{\pm}7$
& {$0.70{\pm}.04$} & {$37{\pm}6$} \\

SQRL~\cite{thananjeyan2021recovery}
& $0.75{\pm}.02$ & $220{\pm}5$
& $0.61{\pm}.01$ & $358{\pm}5$
& {$0.68{\pm}.04$} & {$35{\pm}3$} \\

SEVI(w/o Unc.)
& $0.77{\pm}.03$ & $234{\pm}5$
& $0.63{\pm}.03$ & $360{\pm}7$
& {$0.66{\pm}.04$} & {$32{\pm}5$} \\

\midrule
\textbf{SEVI}
& $\mathbf{0.81{\pm}.02}$ & $\mathbf{154{\pm}3}$
& $\mathbf{0.67{\pm}.01}$ & $\mathbf{310{\pm}5}$
& {$\mathbf{0.83{\pm}.03}$} & {$\mathbf{15{\pm}4}$} \\

\bottomrule
\end{tabular}
}
\vspace{-2mm}
\end{table}

\section{Conclusion}
We presented Safe Evidential Real-World Reinforcement Learning (SEVI), a modular framework that decouples task-policy learning from execution-time safety intervention. SEVI combines an Evidential Safety Critic, which estimates future constraint-violation risk and epistemic uncertainty in a single forward pass, with an offline-trained recovery policy that intervenes when proposed actions are predicted to be unsafe or insufficiently supported by evidence. Across two simulated manipulation tasks and a real-world USB insertion task, SEVI improves safety prediction under distribution shift and reduces unsafe interactions while maintaining task performance. {The current framework has several practical limitations. Recovery behavior is not explicitly optimized for local or motion-efficient correction, and repeated task--recovery switching may occur near failure boundaries during early stages of policy learning. In addition, the learned safety components do not provide formal guarantees of zero violations. Future work will focus on more efficient corrective behaviors, improved switching mechanisms, and tighter integration with formally constrained safety methods.}

\bibliographystyle{IEEEtran}
\bibliography{references}

\clearpage

\twocolumn[{
\begin{center}
{\LARGE \textbf{SEVI: Safe Evidential Real-World Reinforcement Learning}\par}
\vspace{0.25em}
{\LARGE \textbf{Supplementary Material}\par}
\vspace{1.0em}
\end{center}
}]

The supplementary material provides additional details supporting the experimental protocol, implementation, and reproducibility of SEVI. Sec.~\ref{supp:safety-labels} describes task-specific success criteria and safety-label construction. Sec.~\ref{supp:ood-details} details the in-distribution and out-of-distribution evaluation setup. Sec.~\ref{supp:threshold-calibration} explains threshold calibration and sensitivity analysis. Sec.~\ref{supp:baseline-ablation} provides baseline and ablation implementation details, and Sec.~\ref{supp:implementation-details} reports additional implementation details and hyperparameters. Sec.~\ref{sec:supp_metrics} defines the evaluation metrics used in the main manuscript. Sec.~\ref{sec:supp_coverage} provides the training-data coverage analysis of epistemic uncertainty. Sec.~\ref{sec:supp_multirun} reports the repeated hardware evaluation, and Sec.~\ref{sec:supp_recovery_behavior} details the additional recovery-behavior evaluation.

\section{Safety Labels and Success Criteria}
\label{supp:safety-labels}

Safety supervision is derived from trajectory outcomes rather than oracle labels for every state--action pair. In simulation, failures are detected using privileged task signals such as object poses, object drops, workspace violations, and insertion failures. In real-world experiments, trajectories are manually annotated by marking the first failure timestep. Once a failure occurs, all subsequent timesteps are labeled unsafe for safety-critic training, allowing the critic to learn future-risk targets through temporal propagation.

For \textbf{Block Plucking}, success is achieved when the middle yellow block is removed from the three-block stack, the top red block remains supported by the bottom blue block, and the extracted yellow block is not dropped and is at least $7$~cm away from the blue block in the $xy$-plane. A safety violation is labeled if the yellow or red block is dropped, or if the red block loses support or is displaced by more than $10$~cm from the blue block in the $xy$-plane. For \textbf{Transport Nut}, success requires transporting the square nut and inserting it onto the target peg. Safety violations include nut drops, workspace departures, and insertion jamming due to large pose misalignment. For \textbf{USB Insertion}, success is determined using TCP pose thresholds of $2$~mm in $x,y$, $5$~mm in $z$, and $0.15$~rad in orientation, optionally combined with a vision-based reward classifier. Safety violations are defined by keep-out-region breaches and excessive external contact force.

For evaluation, unsafe events are counted separately from timestep-level training labels. A continuous failure segment may contribute multiple unsafe-labeled timesteps to the safety-critic dataset, but it is counted as one unsafe event unless the robot is reset and a new failure occurs.
\section{OOD Evaluation Details}
\label{supp:ood-details}

We evaluate distribution shift separately from closed-loop task learning to test whether the safety critic remains reliable when observations differ from the training distribution. In simulation, OOD rollouts are generated by perturbing both observation and task-configuration factors outside the nominal training distribution. For Block Plucking, we vary camera viewpoints and shift the block-stack configuration, including stack pose, block alignment, and near-boundary extraction configurations that are underrepresented in the training data. For Transport Nut, we perturb the nut and peg poses to induce grasping and insertion misalignment.

For hardware evaluation, we create OOD conditions in the USB insertion task by introducing novel distractor objects, changing cable configurations, and modifying the visual layout around the insertion region. Across all tasks, these perturbations preserve the task objective and safety definitions while changing the visual and geometric context observed by the safety critic. To quantify the magnitude of the shift, we compute the Mahalanobis distance in the latent feature space of the safety critic. Let $z$ denote the critic latent feature before the evidential output head, and let $\mu_{\mathrm{ID}}$ and $\Sigma_{\mathrm{ID}}$ denote the mean and covariance estimated from in-distribution validation rollouts.The distance is computed as
\begin{equation}
M(z)=
\sqrt{
(z-\mu_{\mathrm{ID}})^{\top}
\Sigma_{\mathrm{ID}}^{-1}
(z-\mu_{\mathrm{ID}})
}.
\label{eq:mahalanobis_distance}
\end{equation}

For the USB task, OOD distractor and cable-variation rollouts yield $M>8.5$, whereas in-distribution rollouts have an average distance of approximately $M\approx1.2$. This confirms that the hardware OOD setting induces a substantial feature-space shift relative to the nominal training distribution. For safety-prediction and uncertainty-quality analysis, we collect separate in-distribution and out-of-distribution evaluation rollouts for each task. Simulation rollouts follow the 300-step horizon used in the MuJoCo tasks, while USB insertion rollouts follow the 120-step hardware horizon. We report ID metrics to evaluate prediction reliability under nominal conditions and OOD metrics to evaluate sensitivity to shifted or poorly supported state--action regions.

\begin{table}[t]
\centering
\caption{Task-specific recovery-gating thresholds and sensitivity to $\pm20\%$ threshold perturbation.}
\label{tab:supp_thresholds}
\renewcommand{\arraystretch}{1.08}
\small
\begin{tabular}{l|ccc}
\hline
\textbf{Task} 
& $\boldsymbol{\tau_{\mathrm{risk}}}$ 
& $\boldsymbol{\tau_{\mathrm{epi}}}$ 
& $\boldsymbol{\Delta}$ \textbf{Success} \\
\hline
Block Plucking & $0.44$ & $0.56$ & $<8\%$ \\
Transport Nut & $0.61$ & $0.64$ & $<9\%$ \\
USB Insertion & $0.65$ & $0.45$ & $<5\%$ \\
\hline
\end{tabular}
\end{table}

\section{Threshold Calibration and Sensitivity}
\label{supp:threshold-calibration}

For each task, the risk threshold $\tau_{\mathrm{risk}}$ and epistemic uncertainty threshold $\tau_{\mathrm{epi}}$ are calibrated once before online RL using a held-out validation set containing $50$ successful and $50$ failure rollouts. This validation set is separate from the final evaluation rollouts. We sweep candidate threshold pairs and select the operating point that best separates safe and unsafe validation actions while avoiding degenerate behavior in which nearly all actions are either accepted or rejected. The calibrated thresholds are then kept fixed throughout online training and evaluation.

Table~\ref{tab:supp_thresholds} reports the calibrated thresholds for each task. To evaluate sensitivity, we perturb both thresholds by $\pm20\%$ around the calibrated values and repeat the closed-loop evaluation. The final success rate changes by less than $8\%$ on Block Plucking, less than $9\%$ on Transport Nut, and less than $5\%$ on USB Insertion. These results indicate that SEVI is reasonably robust to threshold variation and does not rely on a narrow threshold choice.

\section{Baseline and Ablation Details}
\label{supp:baseline-ablation}
All task-learning baselines use the same task observations, action spaces, reward definitions, and online interaction budgets as SEVI. SERL and SAC are evaluated without execution-time recovery filtering, while SAC-C uses the same task-learning setup with an additional constraint-penalty term. The SQRL-style recovery baseline uses a point-estimate safety critic trained with the same future-violation risk objective, but shares the offline-trained recovery policy used by SEVI.

To isolate the effect of the safety critic and recovery-gating rule, all recovery-based variants use the same recovery controller. Specifically, SQRL-style recovery, SEVI without uncertainty, and full SEVI differ only in how the proposed task action is evaluated before execution. SQRL-style recovery uses a point-estimate risk critic, SEVI without uncertainty uses the evidential critic but gates actions only by predicted mean risk, and full SEVI gates actions using both predicted risk and epistemic uncertainty.

For standalone safety-critic evaluation, we compare ESC against a point-estimate SQRL-style critic, MC Dropout, and Deep Ensembles using comparable observation--action inputs, safety-label supervision, and training budgets. MC Dropout uses $T=10$ stochastic forward passes per action with dropout enabled at inference time. Deep Ensembles use $M=5$ independently initialized safety critics, and uncertainty is estimated from disagreement across their predictions. In contrast, ESC predicts both future violation risk and epistemic uncertainty in a single forward pass, which enables lower-latency execution-time safety filtering.

\section{Implementation Details}
\label{supp:implementation-details}
This section provides additional implementation details for reproducing SEVI. 
\vspace{-10pt}

\subsection{Evidential Safety Critic}
The Evidential Safety Critic (ESC) is implemented as a twin-critic architecture adapted for safety-risk prediction. Each critic takes the observation--action pair $(o_t,a_t)$ as input and outputs the Normal--Inverse--Gamma parameters $(m_t,\nu_t,\alpha_t,\beta_t)$, which parameterize the predictive distribution over the latent safety value and its uncertainty. For visual tasks, front and wrist RGB observations are encoded using a frozen ResNet-v1-10 encoder with shared weights. The resulting visual embeddings are concatenated with projected proprioceptive and action features and passed to the evidential critic head. For target computation and action filtering, we conservatively aggregate the two critics by taking the maximum predicted risk and maximum epistemic uncertainty. The main ESC hyperparameters are summarized in Table~\ref{tab:supp_esc_hyperparams}.

\begin{table}[t]
\centering
\caption{Evidential Safety Critic hyperparameters.}
\label{tab:supp_esc_hyperparams}
\renewcommand{\arraystretch}{1.08}
\small
\begin{tabular}{l|c}
\hline
\textbf{Hyperparameter} & \textbf{Value} \\
\hline
Critic architecture & Twin critic \\
Visual encoder & Frozen ResNet-v1-10 \\
Camera views & Front + wrist RGB \\
Hidden layers & $[256,256]$ \\
Activation & Swish \\
Normalization & LayerNorm \\
Output parameters & $(m,\nu,\alpha,\beta)$ \\
Optimizer & Adam \\
Learning rate & $3\times10^{-4}$ \\
Batch size & $256$ \\
Risk discount $\gamma_{\mathrm{risk}}$ & $0.99$ \\
Target network $\tau$ & $0.005$ \\
Target clipping $\epsilon$ & $10^{-2}$ \\
Aggregation & Max over twin critics \\
\hline
\end{tabular}
\end{table}

\subsection{Task Policy Learning}

The task policy $\pi_{\mathrm{task}}$ is trained using the SERL/RLPD off-policy RL pipeline with prior demonstrations and online replay. The task learner is kept fixed across all methods so that performance differences arise from the safety critic, action filtering, and recovery mechanism rather than changes to the underlying RL algorithm. SEVI does not modify the task reward, actor objective, critic update, or replay sampling procedure. The key hyperparameters are summarized in Table~\ref{tab:supp_task_policy_hyperparams}.

\begin{table}[t]
\centering
\caption{Task policy hyperparameters.}
\label{tab:supp_task_policy_hyperparams}
\renewcommand{\arraystretch}{1.08}
\small
\begin{tabular}{l|c}
\hline
\textbf{Hyperparameter} & \textbf{Value} \\
\hline
Algorithm & SERL/RLPD \\
Actor hidden layers & $[256,256]$ \\
Critic hidden layers & $[256,256]$ \\
Actor learning rate & $3\times10^{-4}$ \\
Critic learning rate & $3\times10^{-4}$ \\
Replay buffer size & $200$k transitions \\
Update-to-data ratio & $4$ \\
Entropy tuning & Automatic \\
Target entropy & $-\dim(\mathcal{A})/2$ \\
Prior demonstrations & $50$ successful demos \\
\hline
\end{tabular}
\end{table}

\subsection{Recovery Policy}

The recovery policy $\pi_{\mathrm{rec}}$ is trained offline using TD3+BC on teleoperated recovery trajectories collected near task-specific failure boundaries. The recovery policy is not trained to solve the full task; instead, it learns local corrective behaviors that move the robot from near-failure configurations back toward safer regions where the task policy can resume execution. This local objective has a narrower state--action distribution than full task completion, making it feasible to learn from fewer demonstrations.

The recovery objective combines value-based learning with behavior cloning regularization. The value-based component encourages low-risk corrective actions, while the behavior-cloning term keeps the learned recovery actions close to demonstrated teleoperated behavior, reducing extrapolation in the low-data offline setting. In our implementation, the recovery reward depends only on the predicted safety risk from the ESC; epistemic uncertainty is used to trigger recovery at execution time and is not included as an explicit recovery-reward penalty.

We use $150$ recovery trajectories for each simulation task and $200$ trajectories for the hardware USB insertion task. The same recovery dataset and trained recovery policy are used across all recovery-based variants, so performance differences are attributed to the safety critic and gating rule rather than to different recovery controllers. Table~\ref{tab:supp_recovery_policy_hyperparams} summarizes the recovery-policy hyperparameters.

\begin{table}[t]
\centering
\caption{Recovery policy hyperparameters.}
\label{tab:supp_recovery_policy_hyperparams}
\renewcommand{\arraystretch}{1.08}
\small
\begin{tabular}{l|c}
\hline
\textbf{Hyperparameter} & \textbf{Value} \\
\hline
Algorithm & TD3+BC \\
Actor hidden layers & $[256,256]$ \\
Critic hidden layers & $[256,256]$ \\
Actor learning rate & $1\times10^{-4}$ \\
Critic learning rate & $1\times10^{-4}$ \\
Discount factor $\gamma$ & $0.95$ \\
BC weight $\lambda_{\mathrm{BC}}$ & $2.5$ \\
Recovery demos, simulation & $150$ trajectories/task \\
Recovery demos, hardware & $200$ trajectories \\
\hline
\end{tabular}
\end{table}
\section{Evaluation Metric Definitions}
\label{sec:supp_metrics}

This section provides the detailed definitions of the evaluation metrics used in the main manuscript. For trajectory-level safety prediction, we report the true-positive rate (TPR), true-negative rate (TNR), balanced accuracy (BA), normalized detection time (DT), and inference latency. Let TP, TN, FP, and FN denote the numbers of true positives, true negatives, false positives, and false negatives, respectively. We compute
\begin{equation}
\mathrm{TPR} =
\frac{\mathrm{TP}}{\mathrm{TP}+\mathrm{FN}},
\qquad
\mathrm{TNR} =
\frac{\mathrm{TN}}{\mathrm{TN}+\mathrm{FP}},
\end{equation}
and balanced accuracy as
\begin{equation}
\mathrm{BA} =
\frac{1}{2}
\left(
\mathrm{TPR}+\mathrm{TNR}
\right).
\end{equation}

For runtime failure detection, the normalized detection time is defined as
\begin{equation}
\mathrm{DT} =
\frac{t_{\mathrm{det}}}{H},
\end{equation}
where $t_{\mathrm{det}}$ denotes the first timestep at which the safety critic detects an unsafe trajectory and $H$ is the corresponding episode horizon. This normalization makes detection time comparable across tasks with different rollout lengths. Lower DT therefore indicates earlier detection. Inference latency measures the wall-clock computation time required to evaluate one state--action pair and obtain the safety quantities used for execution-time filtering. For sampling-based uncertainty baselines, this measurement includes all stochastic forward passes or ensemble evaluations required to produce the corresponding uncertainty estimate.

For uncertainty quality, we report the area under the receiver operating characteristic curve (AUROC), average precision (APR), expected calibration error (ECE), and Brier score. AUROC evaluates ranking quality across all decision thresholds, while APR summarizes the precision--recall trade-off and is particularly informative when the evaluated classes are imbalanced. To compute ECE, predictions are partitioned into $B$ confidence bins $\{\mathcal{B}_b\}_{b=1}^{B}$. For each bin, we compare the empirical accuracy with the mean predicted confidence:
\begin{equation}
\mathrm{ECE}
=
\sum_{b=1}^{B}
\frac{|\mathcal{B}_b|}{N}
\left|
\mathrm{acc}(\mathcal{B}_b)
-
\mathrm{conf}(\mathcal{B}_b)
\right|.
\end{equation}
Lower ECE indicates better calibration. For binary outcomes, the Brier score is
\begin{equation}
\mathrm{Brier}
=
\frac{1}{N}
\sum_{i=1}^{N}
(p_i-y_i)^2,
\end{equation}
where $p_i$ is the predicted probability and $y_i\in\{0,1\}$ is the corresponding ground-truth label. Lower Brier score indicates better probabilistic accuracy and calibration. For closed-loop execution, task success rate is computed from the final policy evaluation rollouts. Unsafe events and recovery interventions are counted over the complete online training session, while cycle time measures the elapsed execution time required to complete one task cycle.
\section{Epistemic Uncertainty Under Training-Data Coverage}
\label{sec:supp_coverage}

We further characterize whether the epistemic uncertainty used by SEVI reflects limited support in the training data. This experiment corresponds to Fig.~\ref{fig:esc_characterization}(a) of the main manuscript. We consider a fixed failure-prone region of the real-world USB Insertion task in which the robot enters a narrow configuration between the VGA ports and becomes stuck during execution. We collected 200 failure trajectories covering this region and constructed three ESC training subsets containing $10\%$, $50\%$, and $100\%$ of the available data, corresponding to 20, 100, and 200 trajectories, respectively. A separate set of 50 held-out failure trajectories was collected from the same VGA-port failure region and used for evaluation under all three coverage conditions.

For each training-coverage level, the ESC was trained from scratch using three independent random seeds, resulting in nine independently trained ESCs in total. To preserve the temporal structure of the uncertainty signal, the 50 held-out trajectories were aligned over the same 100-step interval leading to the stuck configuration. At each timestep, we first compute the mean epistemic uncertainty over the 50 held-out trajectories for each independently trained ESC. We then compute the mean and standard deviation across the three independent ESC trainings.

As reported in Fig.~4(a), increasing the amount of training data covering the same state--action region results in a consistent reduction in epistemic uncertainty. The ESC trained with $10\%$ coverage exhibits the highest uncertainty throughout the evaluated interval, the $50\%$ coverage condition produces an intermediate uncertainty profile, and the $100\%$ coverage condition remains consistently lower over the same region. Across the 100-step evaluation interval, the standard deviation across independent ESC trainings ranges approximately from $0.02$--$0.08$ for $10\%$ coverage, $0.01$--$0.10$ for $50\%$ coverage, and $0.02$--$0.06$ for $100\%$ coverage.

Because the evaluated state--action region, held-out trajectories, and temporal alignment are fixed across all conditions, while only the amount of training data covering that region is varied, the observed reduction provides empirical evidence that the epistemic uncertainty in Eq.~(17) reflects limited data support. In contrast, aleatoric uncertainty represents irreducible variability arising from effects such as contact stochasticity or sensor noise. SEVI does not use the aleatoric component for recovery intervention; recovery gating is based on the epistemic uncertainty defined in Eq.~(17).

\section{Repeated Hardware Evaluation}
\label{sec:supp_multirun}

To characterize run-to-run variability in real-world online RL, we repeat the USB Insertion hardware evaluation using three independent training sessions for each evaluated method. Each hardware training session lasts 3 hours, corresponding to a total of 9 hours of online hardware interaction per method. Across the three runs, we keep the task configuration, prior demonstrations, recovery policy, safety critic configuration, gating thresholds, and online interaction budget fixed. For each run, the final policy is evaluated over 50 hardware rollouts. Unsafe events are counted throughout the complete 3-hour online training session. The reported hardware results are therefore summarized as mean$\pm$standard deviation across the three independent runs.

\begin{table}[t]
\centering
\caption{Repeated hardware evaluation on the USB Insertion task. Success is evaluated using the final policy, while unsafe events are counted throughout online training. Results are mean$\pm$standard deviation over three independent 3-hour hardware training runs.}
\label{tab:supp_hardware_runs}
\small
\setlength{\tabcolsep}{6pt}
\begin{tabular}{lcc}
\toprule
Method & Success $\uparrow$ & Unsafe $\downarrow$ \\
\midrule
SERL
& $0.71\pm0.03$
& $56\pm7$ \\

SAC
& $0.61\pm0.03$
& $49\pm4$ \\

SAC-C
& $0.70\pm0.04$
& $37\pm6$ \\

SQRL
& $0.68\pm0.04$
& $35\pm3$ \\

SEVI (w/o Unc.)
& $0.66\pm0.04$
& $32\pm5$ \\

\textbf{SEVI}
& $\mathbf{0.83\pm0.03}$
& $\mathbf{15\pm4}$ \\
\bottomrule
\end{tabular}
\end{table}

As shown in Table~\ref{tab:supp_hardware_runs}, the repeated-run evaluation preserves the conclusion of the closed-loop comparison. SEVI achieves a final task success rate of $0.83\pm0.03$ with $15\pm4$ unsafe events, compared with $0.68\pm0.04$ success and $35\pm3$ unsafe events for the SQRL recovery baseline. This corresponds to a $57.1\%$ reduction in unsafe events relative to SQRL. Across the evaluated methods, SEVI achieves the highest final success rate and the lowest unsafe-event count. These aggregate results indicate that the observed safety--performance improvement is consistent across independent hardware training runs rather than being specific to a single training session. The learning curves in Fig.~6 of the main manuscript are shown only to illustrate the temporal evolution of one representative hardware run; the quantitative hardware comparison in Table~V and the results reported here use the aggregate statistics across all three independent runs.

\section{Additional Recovery-Behavior Evaluation}
\label{sec:supp_recovery_behavior}

To assess whether the benefit of uncertainty-aware gating depends on a particular corrective response, we evaluate two qualitatively distinct recovery behaviors on the real-world USB Insertion task: \emph{Hit-and-Recover}, where the robot disengages from an unintended contact configuration before returning control to the task policy, and \emph{Stuck-and-Recover}, where the robot escapes or repositions from a stalled configuration before task execution resumes. Representative hardware demonstrations are provided in the supplementary video.

To isolate the effect of the gating criterion, we keep the task policy, ESC, calibrated thresholds, recovery mechanism, and hardware setup fixed, and compare risk-only gating against risk-plus-epistemic-uncertainty gating. We perform three independent evaluation runs for each condition and report mean$\pm$standard deviation.

\begin{table}[t]
\centering
\caption{Closed-loop performance under two qualitatively distinct recovery behaviors on the USB Insertion task. Results are mean$\pm$standard deviation over three independent evaluation runs.}
\label{tab:supp_recovery_behavior}
\small
\setlength{\tabcolsep}{5pt}
\begin{tabular}{llcc}
\toprule
Recovery Behavior & Gating & Success $\uparrow$ & Unsafe $\downarrow$ \\
\midrule
Hit-and-Recover
& Risk only
& $0.64\pm0.03$
& $32\pm4$ \\
& Risk + Unc.
& $\mathbf{0.82\pm0.03}$
& $\mathbf{16\pm4}$ \\
\midrule
Stuck-and-Recover
& Risk only
& $0.64\pm0.01$
& $33\pm4$ \\
& Risk + Unc.
& $\mathbf{0.83\pm0.03}$
& $\mathbf{15\pm5}$ \\
\bottomrule
\end{tabular}
\end{table}

As shown in Table~\ref{tab:supp_recovery_behavior}, risk-plus-uncertainty gating improves task success and reduces unsafe events under both recovery behaviors. These results indicate that the advantage of uncertainty-aware gating is not specific to a single corrective motion: the gate determines when recovery is invoked, while the recovery behavior determines how the robot moves away from an unsafe or unreliable configuration.

\vfill

\end{document}